\documentclass[journal]{IEEEtran}
\usepackage{ifpdf}
\usepackage{cite}
\ifCLASSINFOpdf
\usepackage[pdftex]{graphicx}
\graphicspath{{../pdf/}{../jpeg/}}
\DeclareGraphicsExtensions{.pdf,.jpeg,.png}
\else
\usepackage[dvips]{graphicx}
\graphicspath{{../eps/}}
\DeclareGraphicsExtensions{.eps}
\fi
\usepackage{amsmath}
\usepackage{amssymb}
\usepackage{algorithmic}

\usepackage{array}
\usepackage{booktabs}
\usepackage{threeparttable}
\usepackage{multirow}
\newcommand{\RNum}[1]{\uppercase\expandafter{\romannumeral #1\relax}}
\usepackage{tabularx}

\begin{document}
%
% paper title
% Titles are generally capitalized except for words such as a, an, and, as,
% at, but, by, for, in, nor, of, on, or, the, to and up, which are usually
% not capitalized unless they are the first or last word of the title.
% Linebreaks \\ can be used within to get better formatting as desired.
% Do not put math or special symbols in the title.
\title{Shared Control for Bimanual Telesurgery with Optimized Robotic Partner}
%
%
% author names and IEEE memberships
% note positions of commas and nonbreaking spaces ( ~ ) LaTeX will not break
% a structure at a ~ so this keeps an author's name from being broken across
% two lines.
% use \thanks{} to gain access to the first footnote area
% a separate \thanks must be used for each paragraph as LaTeX2e's \thanks
% was not built to handle multiple paragraphs
%
 
\author{Ziwei~Wang,~\IEEEmembership{Member,~IEEE,}
       Yanpei~Huang,~\IEEEmembership{Member,~IEEE,}
       Xiaoxiao~Cheng,~\IEEEmembership{Member,~IEEE,}\\
       Pakorn~Uttayopas,
       Etienne~Burdet,~\IEEEmembership{Member,~IEEE}
        
% <-this % stops a space
\thanks{This research was partly supported by the European Commission grants H2020 PH-CODING (FETOPEN 829186), NIMA (FETOPEN 899626) and UK EPSRC grant FAIR-SPACE EP/R026092/1. (Corresponding author: Yanpei Huang and Xiaoxiao Cheng)} 

\thanks{The authors are with the Department of Bioengineering, Imperial College of Science, Technology and Medicine, SW7 2AZ, United Kingdom (e-mail: \{ziwei.wang, yanpei.huang, xiaoxiao.cheng,pakorn.uttayopas18, e.burdet\}@imperial.ac.uk).}

}% <-this % stops a space
\maketitle

% As a general rule, do not put math, special symbols or citations
% in the abstract or keywords.
\begin{abstract}
Traditional telesurgery relies on the surgeon's full control of robot on the patient's side, which tends to increase surgeon fatigue and may reduce the efficiency of the operation. This paper introduces a \textit{Robotic Partner} (RP) to facilitate intuitive bimanual telesurgery, aiming at reducing the surgeon workload and enhancing surgeon-assisted capability. An interval type-2 polynomial fuzzy-model-based learning algorithm is employed to extract expert domain knowledge from surgeons and reflect environmental interaction information. Based on this, a bimanual shared control is developed to interact with the other robot teleoperated by the surgeon, understanding their control and providing assistance. As prior information of environment model is not required, it reduces reliance on force sensors in control design. Experimental results on the DaVinci Surgical System show that the RP could assist peg-transfer tasks and reduce the surgeon's workload by 51\% in force-sensor-free scenarios.
\end{abstract}

% Note that keywords are not normally used for peerreview papers.

% For peer review papers, you can put extra information on the cover
% page as needed:
% \ifCLASSOPTIONpeerreview
% \begin{center} \bfseries EDICS Category: 3-BBND \end{center}
% \fi
%
% For peerreview papers, this IEEEtran command inserts a page break and
% creates the second title. It will be ignored for other modes.
\IEEEpeerreviewmaketitle

\section{Introduction}
% The very first letter is a 2 line initial drop letter followed
% by the rest of the first word in caps.
% 
% form to use if the first word consists of a single letter:
% \IEEEPARstart{A}{demo} file is ....
% 
% form to use if you need the single drop letter followed by
% normal text (unknown if ever used by the IEEE):
% \IEEEPARstart{A}{}demo file is ....
% 
% Some journals put the first two words in caps:
% \IEEEPARstart{T}{his demo} file is ....
% 
% Here we have the typical use of a "T" for an initial drop letter
% and "HIS" in caps to complete the first word.
% \IEEEPARstart{C}{urrent} \IEEEPARstart{C}{urrent} telesurgical robots are typically commanded in master-slave scheme.

% Compared to conventional open surgery, minimally invasive surgery (MIS) access to the surgical site through small keyhole incisions, which offers many benefits to patient such as reduced trauma, less blood loss and fast recovery \cite{2009review}. However, manipulating these MIS laparoscopic long instruments is complex and skill-demanding for the surgeons. 

\IEEEPARstart{R}{obotic} technology and its integration with surgical procedures have largely improved the operation intuitiveness for the surgeons \cite{2012MIS}. Most of the current surgical robotic systems are conducted in master-slave mode, in which the surgeon could comfortably sit at a master console, remotely controlling the slave robotic arm and holding surgical tools through master interfaces and high-definition camera view \cite{2021huang,Treratanakulchai2021Passive}. These teleoperation systems tend to simplify the surgeon operation to some extent. However, the surgeon's performance would suffer from fatigue during long time operation. Also, the surgeon's performance is limited by delayed visual and tactile information from the interaction environment due to the communication channel. Therefore, sharing the surgical robot with a degree of autonomy could facilitate the telesurgery, easing both physical and mental workloads for surgeons at the master console. 

To this end, the slave surgical robot needs to perceive interaction forces in terms of telepresence. Nevertheless, it is challenging to obtain interactive forces in contact with patient tissue directly due to the mounting position of the force sensors conflicting with the surgical tool. Some studies \cite{Yang2021Force,Xu2021Force-reflecting} integrated the environment information into the control design by force sensors or model calculation to enable the robot to conduct force interaction with environment automatically. These methods require accurate model structures of remote interactive dynamics or force sensors on the end-effector of the robot. It is revealed in \cite{Wang2020Prescribed-Time} that accurate dynamics model is difficult to build in practical teleoperation subject to uncertainties, which may limit the merits of the above methods in telesurgery.

Another critical issue to realise the target is how the surgical robot can predict the human intention and give assistance, thus reducing the workload. The current methods mostly focus on training the robot to carry out repetitive tasks \cite{Fontanelli2018,Ishida2020}, which lack the control flexibility to reflect on different scenarios. To assist the surgeon better, the surgical robot is required to adapt to the cooperated surgeon, react to the interacted environment and even improve the performance of the surgeon through sensory augmentation \cite{takagi2017physically}. 

Based on this idea, we propose an intelligent \textit{Robotic Partner} (RP) in bimanual shared control scheme to solve the two issues mentioned above. The RP is trained by interval type-2 polynomial fuzzy-model-based learning (IT2 PFMBL) algorithm according to human-human demonstration and robot-environment interaction, which enables the learning speed with few-shot data through virtue of the fitting ability of polynomials and the nonlinear expression ability of fuzzy models. The RP could then estimate interaction force without the exact physical model nor force sensors. Moreover, the proposed RP provides intuitive assistance to perform a cooperation task (e.g. peg-transfer) with human operator.

% is to integrate the environment information into the control design by force sensors or model calculation. Although several works focus on estimating human-environment interaction forces through force observer designs \cite{Yang2021Force,Xu2021Force-reflecting}, 

% The RP could cooperate closely with the surgeon, adapt to the surgeon's specific sensorimotor abilities and perform the task automatically based interaction environment without sensors. 

% To this end, G. Ganesh et al. showed the subjects can improve their performance by interacting with (even a worse) partner in a tracking task \cite{ganesh2014two}. Similar performances with a robotic partner obtained in \cite{takagi2017physically} demonstrated that this property results from the sensory exchange between the two agents via the haptic channel.

% Motivated by the above analysis, the Robotic Partner (RP) trained by interval type-2 polynomial fuzzy-model-based learning (IT2 PFMBL) algorithm is developed to facilitate intuitive surgeon-assistance and interactive force estimation without the need of force sensors. By virtue of the fitting ability of polynomials and the nonlinear expression ability of fuzzy models, the interactive force estimation is data-driven without the need to establish an exact physical model. Only few-shot data is required for interaction identification and demonstration learning. 

% You must have at least 2 lines in the paragraph with the drop letter
% (should never be an issue)

\section{Method}\label{sec2}
IT2 PFMBL algorithm is designed to estimate the human motions (i.e., movement trajectories and gesture actions) and interactive force with environment simultaneously based on the human-robot and robot-environment interaction data. 
% cx:  how could we link it better to the learning of human behaviour especially motion trajectories?  

%\textbf{Step 1}. 
Let collected input-output pair data be $[x_i(k), v_i(k), y_i(k)]_{k=1}^N$, where $i\in\{h_{mt},h_{ga},e\}$ stands for human movement trajectories, gesture actions or interactive forces with environment, respectively. $x_i(k),v_i(k), y_i(k) \in\mathbb{R}^{n_i}$ are position, velocity and the associated output in Cartesian space at the $k$th sampling. $N$ is the number of sampling pairs. Fuzzy C-Means clustering \cite{bezdek2013pattern} and subtractive clustering \cite{chiu1994fuzzy} are utilised to obtain the number of fuzzy sets $a_i$, the centre of each fuzzy set ${\mathcal{B}}_{\nu}^i$ and the corresponding membership function ${\mu}_{\mathcal{B}_{\nu}^i}\mathcal{A}_i^{\nu}(z_i(k))$, where $\nu\in \mathbb{N}_{+}{[1,a]}$, $z_i(k)=[x_i(k)\; v_i(k)\; v_i(k+1)]\in\mathbb{R}^{3n_i}$ and $\mathcal{A}_i^{\nu}(z_i(k))$ is the premise variable. The membership function $0\le\underline{\mu}_{\mathcal{B}_{\nu}^i}\mathcal{A}_i^{\nu}(z_i(k))\le{\mu}_{\mathcal{B}_{\nu}^i}\mathcal{A}_i^{\nu}(z_i(k))\le\bar{\mu}_{\mathcal{B}_{\nu}^i}\mathcal{A}_i^{\nu}(z_i(k))\le1$ describes the degree to which the input data matches the fuzzy set.  

%\textbf{Step 2}.
With the help of the obtained fuzzy sets, the inner relationship between the input and output can be represented by the following polynomial discrete-time fuzzy logic system 
\begin{equation}
  \begin{aligned}
  \text{Rule}\;l:\; &\text{IF}\;\mathcal{A}_i^{1}(z_i(k))\;\text{is}\;	{\mathcal{B}}_{1}^{il}\;\text{AND},...,\text{AND}\;\mathcal{A}_i^{a}(z_i(k))\;\text{is}\;{\mathcal{B}}_{a}^{il}\\
  &\text{THEN}\;y_{i}(k)=M_i^l(z_i(k))\frac{v_i(k+1)-v_i(k)}{\Delta T}\\
  &+C_i^l(z_i(k))v_i(k)+K_i^l(z_i(k)) x_i(k)+f_i^l(z_i(k)),\\
  \end{aligned}
  \label{eq1}
  \end{equation}
  where ${\Delta T}$ is the sampling time; ${\mathcal{B}}_{\nu}^{il}$ is a polynomial fuzzy set of Rule $l$ corresponding to the premise variable $\mathcal{A}_i^{\nu}(z_i(k))$, $l\in \mathbb{N}_{+}{[1,p]}$; $M_i^l(z_i(k))\in\mathbb{R}^{n_i\times n_i}$, $C_i^l(z_i(k)) \in\mathbb{R}^{n_i\times n_i}$, $K_i^l(z_i(k))\in\mathbb{R}^{n_i\times n_i}$ and $f_i^l(z_i(k))\in\mathbb{R}^{n_i}$ are the polynomial system matrices, where each element is a polynomial in $z_i(k)$. For example, in the case of $i=e$, the first three system matrices donate the equivalent inertia, damping and stiffness ones while the last the bounded exogenous force concerned with the position of the contact point. Robust linear least-square approach is employed in each Rule to identify the polynomial coefficients of the associated system matrices. 
  
  %\textbf{Step 3}. 
  Considering the measurement noise in $z_i(k)$ and the system uncertainty, the firing strength of the Rule $l$ is of the following interval set
  \begin{equation}
  \label{eq2}
  W_i^{l}(z_i(k))=[	\underline{\omega}_i^{l}(z_i(k)), \bar{\omega}_i^{l}(z_i(k))], l\in \mathbb{N}_{+}{[1,p]},
  \end{equation}
  where $\underline{\omega}_i^{l}(z_i(k))=\prod_{\nu=1}^{a} \underline{\mu}_{\mathcal{B}_{\nu}^{il}}\mathcal{A}_i^{\nu}(z_i(k))$, $\bar{\omega}_i^{l}(z_i(k))=\prod_{\nu=1}^a \bar{\mu}_{\mathcal{B}_{\nu}^{il}}\mathcal{A}_i^{\nu}(z_i(k))$, in which $\underline{\mu}_{\mathcal{B}_{\nu}^{il}}\mathcal{A}_i^{\nu}(z_i(k))$ and $\bar{\mu}_{\mathcal{B}_{\nu}^{il}}\mathcal{A}_i^{\nu}(z_i(k))$ denote the lower and upper membership function such that $0\le\underline{\mu}_{\mathcal{B}_{\nu}^{il}}\mathcal{A}_i^{\nu}(z_i(k)) \le \bar{\mu}_{\mathcal{B}_{\nu}^{il}}\mathcal{A}_i^{\nu}(z_i(k)) \le 1$; $\underline{\omega}_i^{l}(z_i(k))$ and $\bar{\omega}_i^{l}(z_i(k))$ are the lower and upper grade of membership, respectively, satisfying $0\le\underline{\omega}_i^{l}(z_i(k))\le\bar{\omega}_i^{l}(z_i(k))$. Hence, the data-driven IT2 PFMBL system is described by
  \begin{equation}
  \begin{aligned}
  \label{eq3}
  y_{i}(k)=&\sum_{l=1}^p\tilde{\omega}_i^{l}(z_i(k))\Bigg(M_i^{l}(z_i(k))\frac{v_i(k+1)-v_i(k)}{\Delta T}\\
  &+C_i^{l}(z_i(k))v_i(k)+K_i^{l}(z_i(k)) x_i(k)+f_i^{l}(z_i(k))\Bigg),\\
  \end{aligned}
  \end{equation}
  and
  \begin{equation}
  \label{eq4}
  \tilde{\omega}_i^{l}(z_i(k))=\underline{\omega}_i^{l}(z(k))\underline{b}_i^{l}(z_i(k))+\bar{\omega}_{l}(z_i(k))\bar{b}_i^{l}(z_i(k)),
  \end{equation}
  with $\underline{b}_i^{l}(z_i(k))$, $\bar{b}_i^{l}(z_i(k))\in [0,1]$ such that $\underline{b}_i^{l}(z_i(k))+\bar{b}_i^{l}(z_i(k))=1$.
  
\textit{Remark 1}. A unified data-driven learning paradigm is developed without the prior information (i.e., deterministic model structure) in  \cite{Shahdi2009Adaptive,Lu2020High-gain}. For $i\in\{h_{mt},h_{ga},e\}$, $y_i$ represents the human motion trajectories in Cartesian space, gripper angles and interactive forces with the environment, leading to different system matrices for different tasks.

\section{Experimental validation and results}
In this section, we have verified the proposed method by the following experiments, which provides implementation foundation for the RP.
%The proposed IT2 PFMBL method provides an effective way to train the RP in a short training time with few-shot demos. The RP then could interact with environment and cooperate with surgeon to perform surgical operation including contact force identification and motion trajectories generation.

% different operations in multiple scenarios. In this section, we have verified the proposed method in two common surgical operation situations of environmental force estimation and shared control with human operator.  

\subsection{Prediction of robot-environment interaction force}
\begin{figure}[htbp]
\centering
\includegraphics[width=0.4\textwidth]{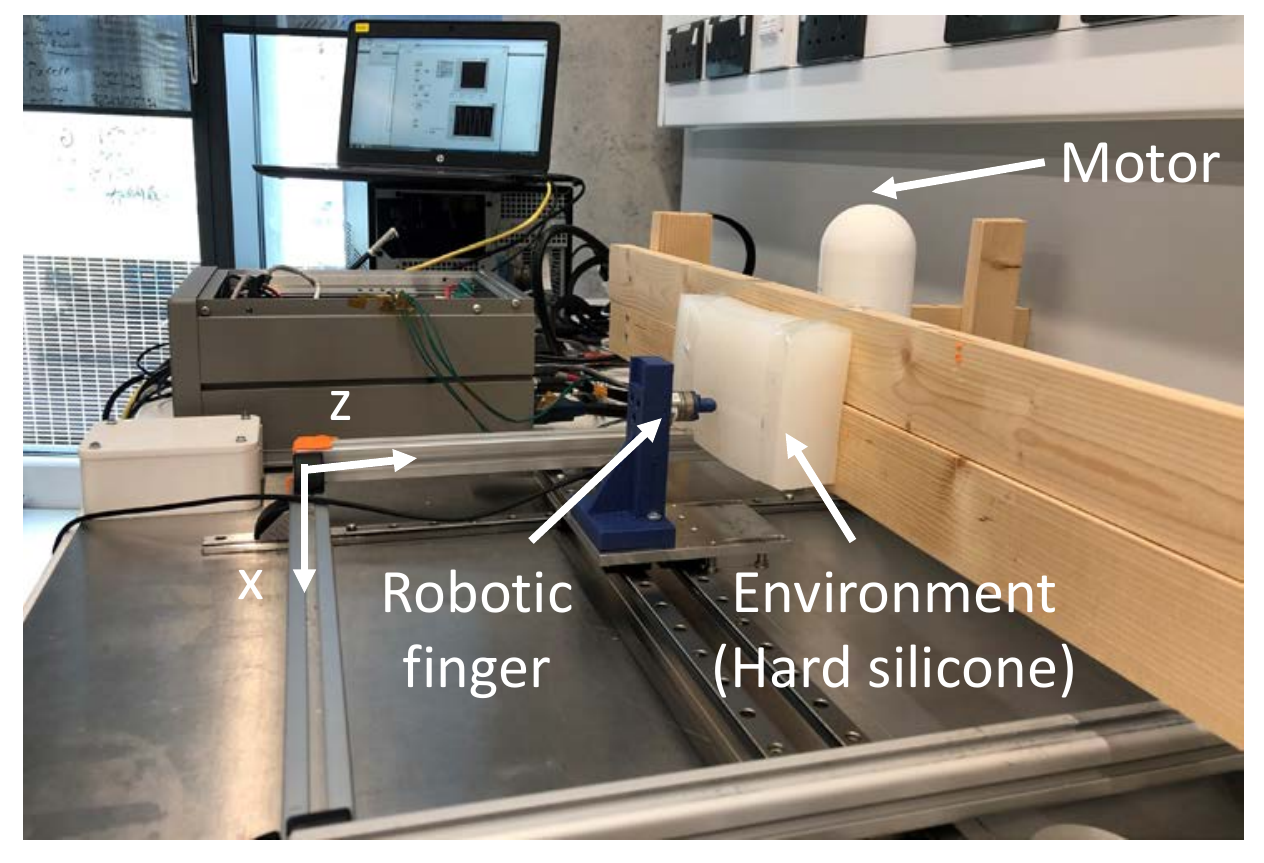}
\caption{H-Man planar robot-environment interaction platform with robotic finger.} 
\label{H-man_setup}
\end{figure}

\begin{figure}[htbp]
\centering
\includegraphics[width=0.5\textwidth]{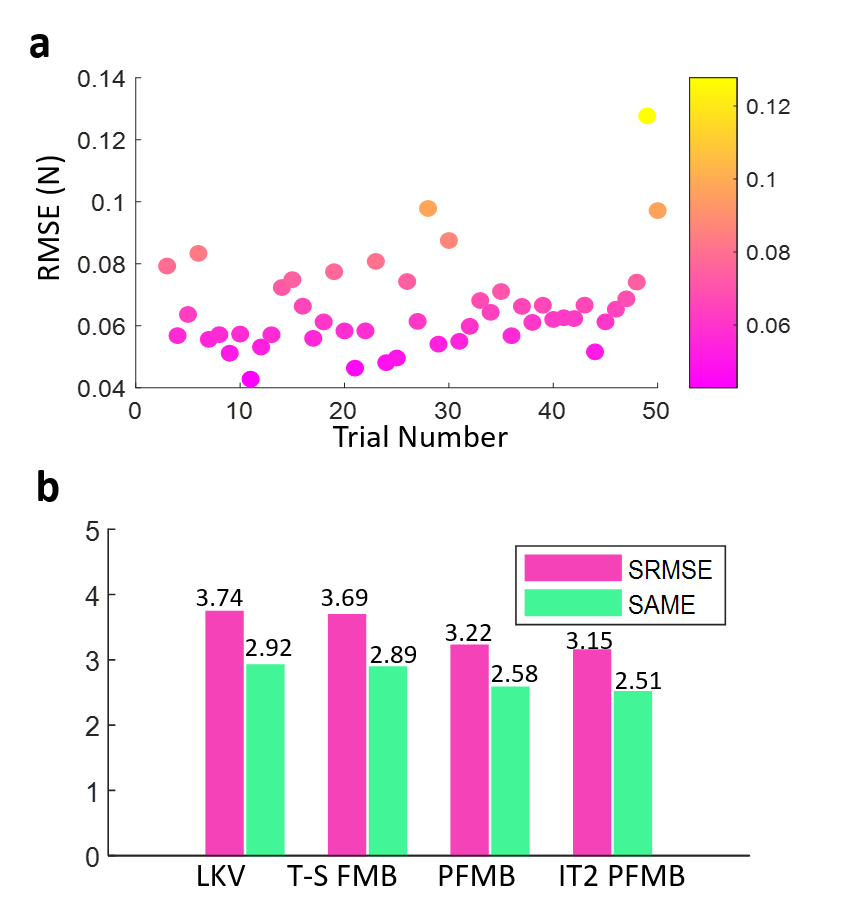}
\caption{(a) RMSE of interactive force estimation using IT2 PFMBL algorithm. (b) Performance comparison in terms of SRMSE and SAME (The smaller these two indicators are, the better).} 
\label{H-man_result}
\end{figure}

We firstly apply the proposed IT2 PFMBL method to predict the interaction force in a planar robot-environment interaction platform using modified H-Man system \cite{CAMPOLO2014H-Man}, which is shown in Fig.\,\ref{H-man_setup}. The environment is selected as hard silicone having similar mechanical characteristics with the organ tissues. The robotic finger equipped with an ATI 6-axis force/torque sensor will press the silicon at three different levels of depth: shallow, medium and deep, and scan the whole environment surface to precept the overall material properties. Interaction force is collected during 50 interactive trials for each type of pressing, totalling 150 trials for each sampling point. The proposed IT2 PFMBL method is advantageous at producing high-prediction accuracy with a small training set by introducing a type-2 fuzzy rule. We randomly allocate available data into 10/90 proportions of training and testing sets. Fig.\,\ref{H-man_result}a shows that the root-mean-square-error (RMSE) of the IT2 PFMBL method decreases rapidly with few-shot training data.  

\begin{figure*}[htbp]
\centering
\includegraphics[width=1\textwidth]{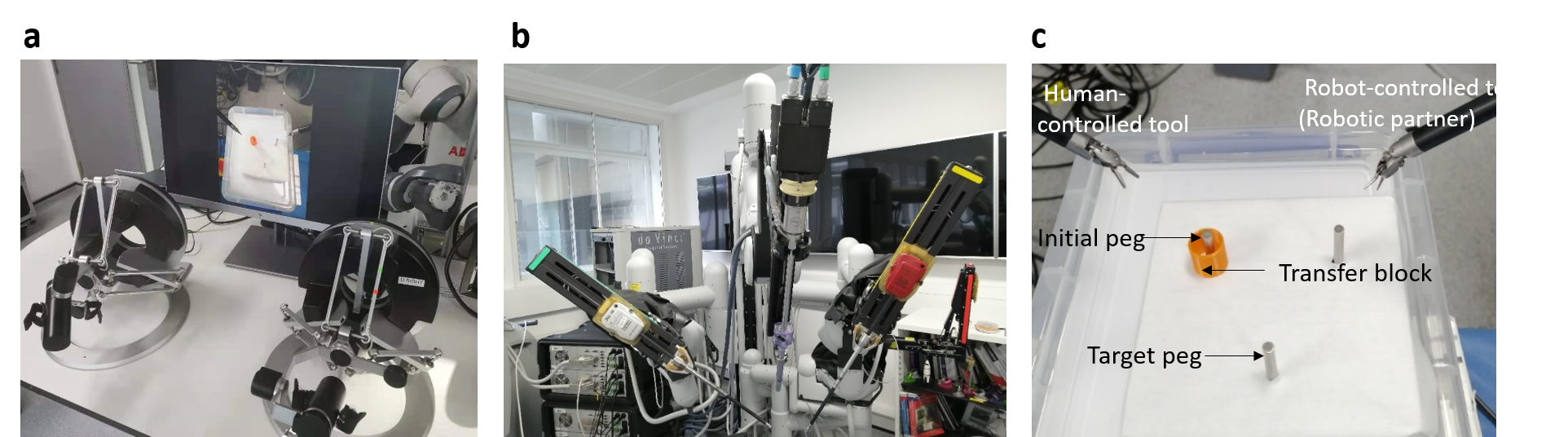}
\caption{Experimental setup. (a) Master interfaces: dual Omega 7 devices with visual feedback. (b) Slave robot: surgical robotic manipulators. (c) Task: peg transfer.}
\label{Davinci_setup}
\end{figure*}

To validate the merits of the IT2 PFMBL method, we compare its performances with other three common interaction force models, namely linear Kelvin-Voigt (LKV) model \cite{Li2014Achieving}, T-S FMB \cite{Wang2020Event-Triggered} and PFMB model. Due to the improved non-linear representation of material stiffness properties, Fig.\,\ref{H-man_result}b indicates the force estimation of IT2 PFMBL surpasses the above models on the sum of RMSE (SRMSE) and mean absolute error (SMAE) in  testing trials.

\subsection{Human-Robot shared control using IT2 PFMBL}
To further assess the performance of the IT2 PFMBL method, we implemented it in a surgical robotic teleoperation system to train the RP to cooperate with a human operator. The setup of the surgical robotic system is shown in Fig.\,\ref{Davinci_setup}. The master console (Fig.\,\ref{Davinci_setup}a) consists of two hand haptic interfaces (Force Dimension Inc., Switzerland) and a monitor providing visual feedback of the patient side. Two patient-side manipulators of Da Vinci robotic system (Intuitive Surgical, USA) are used as slave robotic arms as shown in Fig.\,\ref{Davinci_setup}b.

The end-effectors of the robotic arms are connected to surgical grippers to perform a peg-transfer task (Fig.\,\ref{Davinci_setup}c), which is a standard laparoscopic training task from Fundamentals of Laparoscopic Surgery (FLS) \cite{2011FLS}. The left gripper is controlled by a human operator and the right gripper by a RP. During the task, the human operator navigates the left gripper to the initial peg with an orange block, grasps the block and moves it to conduct a handover to the RP controlled gripper. After the handover, the RP moves the block to the target peg  (Fig.\,\ref{Davinci_setup}c) (see supplementary video 1). The manipulation data of two operators were used as the input to the IT2 PFMBL method, resulting in a final interaction environment model including the motion trajectories, and gripper movements. After training, the RP could imitate the human partner and cooperate intelligently with another human operator. 

% Part of this paragraph may be put into the first paragraph. 

\begin{figure*}[htbp]
\centering
\includegraphics[width=0.9\textwidth]{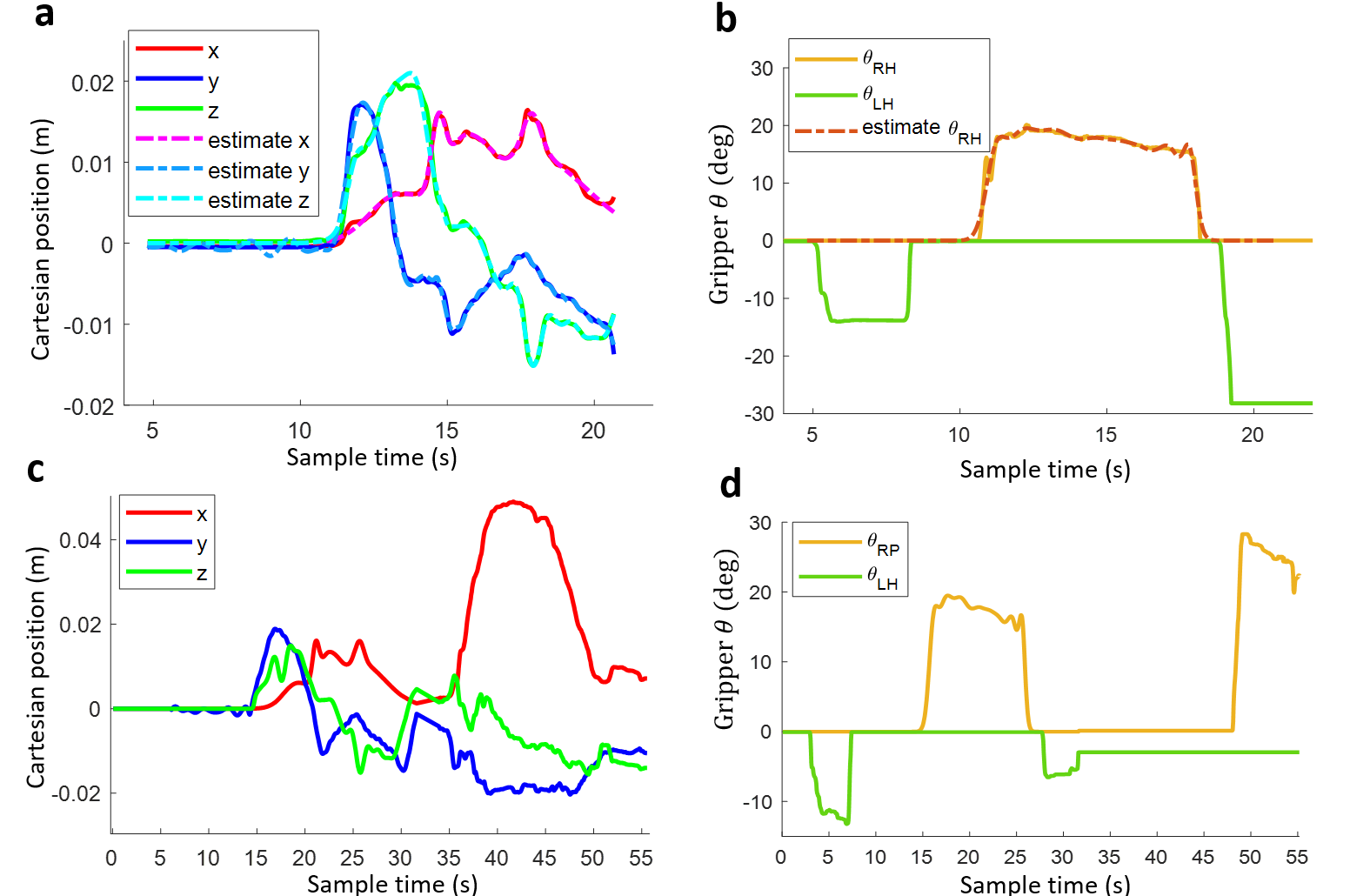}
\caption{Experimental results from (a-b) human-human training to (c-d) human-robot testing. Motion learning from human operator and human partner:(a) Cartesian trajectories in human-human telesurgery (x, y and z denote the right hand movement along the associated direction; the dashed lines are the estimate with respect to the associated direction using IT2 PFMBL). (b) Gripper grasping $\theta$: the dashed line stands for the estimate of gripper movement. $\theta_{\text{LH}}$ and $\theta_{\text{RH}}$ represent the angle at which the left and right hand grippers open. Motion learning from human operator and RP: (c) Cartesian trajectories in human-robot telesurgery (x, y and z denote the RP movement along the associated direction). (d) Assistive gripper grasping: $\theta_{\text{RP}}$ is the angle of RP gripper.
}
\label{Davinci_result}
\end{figure*}

Fig.\,\ref{Davinci_result} a\&b show the comparison of motion trajectories between human-human manipulation and our training results from IT2 PFMBL method. It indicates that the derived models could effectively represent the human partner's behavior with minimal errors. We implemented these models in the surgical robotic system to validate their performance. The results (Fig.\,\ref{Davinci_result}c,d) show that the trained RP could successfully complete the transfer task with closely cooperating with a human partner (see supplementary video 2). The task completion of human operator with RP is within 27 seconds.

%%*************************************************
\section{Discussion and Conclusion}\label{sec4}
An innovative \textit{Robotic Partner} concept is proposed in shared control for bimanual telesurgery to reduce the workload of the surgeon. A novel interval type-2 polynomial fuzzy model-based method is designed to realize precise estimation of contact force and few-shot training for intelligent surgeon assistance. The experimental results illustrate that the proposed IT2 PFMBL method achieves superior nonlinear representation capability and fitting performance in terms of SRMSE and SAME. Compared to T-S FMB method, PFMBL method can better characterise the nonlinearities of interaction force because its model parameters draw on the ability to fit polynomials. Besides, IT2 fuzzy sets are effective to capture the noise and system uncertainty by using the upper and lower bound of membership functions, thereby demonstrating satisfactory estimation robustness. The IT2 PFMBL is therefore able to estimate force accurately and provide haptic feedback to telesurgery system, avoiding installing force sensors on delicate surgical tools. 

Furthermore, the experiment results showed that the RP could reduce the intensity of the surgeon's manipulation effectively with the implemented IT2 PFMBL method, by assisting and adapting to surgeon behaviour autonomously. Trained by trajectories and gripper movements of two human operators, the intuitive human-robot cooperation is demonstrated in a peg-transfer task with self-determining movements from the RP. With the assistance of the RP, the surgeon could teleoperate the other surgical robot in the peg-transfer task alone avoiding the communication errors of two surgeons and saving the manpower. 

\ifCLASSOPTIONcaptionsoff
  \newpage
\fi

% trigger a \newpage just before the given reference
% number - used to balance the columns on the last page
% adjust value as needed - may need to be readjusted if
% the document is modified later
%\IEEEtriggeratref{8}
% The "triggered" command can be changed if desired:
%\IEEEtriggercmd{\enlargethispage{-5in}}

% references section

% can use a bibliography generated by BibTeX as a .bbl file
% BibTeX documentation can be easily obtained at:
% http://mirror.ctan.org/biblio/bibtex/contrib/doc/
% The IEEEtran BibTeX style support page is at:
% http://www.michaelshell.org/tex/ieeetran/bibtex/
\bibliographystyle{IEEEtran}
% argument is your BibTeX string definitions and bibliography database(s)
\bibliography{IEEEabrv,Bibliography}
%
% <OR> manually copy in the resultant .bbl file
% set second argument of \begin to the number of references
% (used to reserve space for the reference number labels box)
%\begin{thebibliography}{1}

%\bibitem{IEEEhowto:kopka}
%H.~Kopka and P.~W. Daly, \emph{A Guide to \LaTeX}, %3rd~ed.\hskip 1em plus
  %0.5em minus 0.4em\relax Harlow, England: Addison-Wesley, 1999.

%\end{thebibliography}

% biography section
% 
% If you have an EPS/PDF photo (graphicx package needed) extra braces are
% needed around the contents of the optional argument to biography to prevent
% the LaTeX parser from getting confused when it sees the complicated
% \includegraphics command within an optional argument. (You could create
% your own custom macro containing the \includegraphics command to make things
% simpler here.)
%\begin{IEEEbiography}[{\includegraphics[width=1in,height=1.25in,clip,keepaspectratio]{mshell}}]{Michael Shell}
% or if you just want to reserve a space for a photo:

% insert where needed to balance the two columns on the last page with
% biographies
%\newpage

% You can push biographies down or up by placing
% a \vfill before or after them. The appropriate
% use of \vfill depends on what kind of text is
% on the last page and whether or not the columns
% are being equalized.

%\vfill

% Can be used to pull up biographies so that the bottom of the last one
% is flush with the other column.
%\enlargethispage{-5in}

% that's all folks
\end{document}